**Author for correspondence:**
Takehiro Ishikawa
e-mail: tishikawa8@gatech.edu

# Balancing centralized learning and distributed self-organization: a hybrid model for embodied morphogenesis

Takehiro Ishikawa

Georgia Institute of Technology College of Computing, Atlanta, USA

TI, 0009-0004-9722-5886

Background: both embodied intelligence and developmental morphogenesis depend on a division of labour between centralized guidance and distributed material dynamics, but the amount of top-down control needed to steer self-organization remains unclear. Methods: we coupled a compact full-resolution convolutional controller to a differentiable Gray–Scott reaction-diffusion (RD) substrate. The controller observes two fields, U and V, and applies smooth gain-scheduled modulations of the feed and kill parameters (ΔF and ΔK). We compared pure RD, neural network (NN)-dominant and hybrid regimes and evaluated spectral selectivity, convergence and control cost. Results: the hybrid regime achieved 100% strict convergence at approximately 165 steps, whereas pure RD and the NN-dominant baseline did not converge within the same horizon. It matched the substrate's spectral selectivity while using approximately 15× less L1 effort and over 200× less L2 power than the NN-dominant controller. Moderate amplitudes ($A \approx 0.03$–$0.045$) formed a Goldilocks zone with 100% quasi-convergence in 94–96 steps. Conclusions: effective control is best framed as seed then cede: brief, smooth parameter-level nudges place the system in a favourable basin of attraction, after which reaction-diffusion dynamics complete and stabilize the pattern. This provides a quantitative model of morphological computation for controlled self-organization.

## 1. Introduction

Patterns in living systems often emerge from local interactions rather than from a central blueprint alone. Reaction-diffusion (RD) theory, beginning with Turing's [1] account of morphogenesis, formalized how simple local reactions and diffusion can amplify microscopic asymmetries into macroscopic order. The Gray–Scott family of models is especially useful in this context because it is simple enough to analyse yet rich enough to generate spots, rings, stripes, labyrinths and other Turing-like textures across small parameter changes [2–4]

A second line of work, from embodied cognition and morphological computation, argues that intelligence is distributed across controller, body and environment rather than residing solely in a centralized brain [5–7]. In developmental biology, related ideas appear in positional information, bioelectric regulation and multi-scale control of tissue patterning, where cells and tissues can integrate local signals while remaining sensitive to longer-range instructive cues [6,8,9]. Our use of the terms brain-like and cell-like is therefore operational rather than literal: the learned controller plays the role of a centralized policy, while the RD substrate plays the role of a distributed material process whose updates are driven by local interactions. The present study does not claim that this controller is cognitive in a strong philosophical or neurobiological sense. Rather, it introduces a minimal operational model for measuring the division of labour between centralized guidance and distributed self-organization.

This framing matters because both development and engineering face the same design question: how much global control should be imposed on a self-organizing system before that control begins to fight the very physics that makes the system robust? Biological analogies range from bioelectric pattern regulation to molluscan shell formation, where neural or neurosecretory influences can bias local pattern-forming processes without explicitly painting the final texture [6,9]. Related artificial-life work on sensorimotor cellular automata likewise asks how agency and control emerge from local update rules coupled to external guidance [10,11].

What is already known is that Gray–Scott dynamics can produce the relevant classes of patterns, and that neural controllers can optimize spatial objectives. What remains unclear, however, is how to quantify the division of labour between centralized guidance and distributed self-organization. The contribution of the present study is not a new RD model in isolation, nor a new neural architecture in isolation, but an explicit quantitative coupling between them: a full-resolution convolutional controller restricted to modulating the RD parameters F and K, an objective that rewards target-band selectivity while penalizing instability and control cost, and a regime comparison that separates self-organization, centralized control and the hybrid case.

We ask a precise question: when is a learned controller useful as a scaffold rather than as a replacement for self-organization? To answer it, we compare (i) pure RD, (ii) a neural network (NN)-dominant baseline, and (iii) a hybrid regime in which a modest controller acts on a strong Gray–Scott substrate. We evaluate strict convergence and quasi-convergence, spectral selectivity, absolute band power and the L1/L2 cost of the controller's interventions. The resulting picture is that moderate, early, smooth control is useful, whereas both under-control and over-control are harmful.

This broader question has clear application domains. In embodied robotics, brief controller actions can entrain passive or compliant body dynamics rather than micromanaging every degree of freedom [12]. In bioengineering and synthetic morphogenesis, small instructive cues can steer multicellular self-assembly without specifying every local interaction [13]. In self-assembly and analogue hardware, the same principle appears when a weak template or control signal nudges a physical system into a productive attractor and then lets the substrate do the computational work [14,15].

# 2. Methods

## 2.1. Code and data availability

The accompanying notebook and archived publication package are available through the Zenodo release (doi:10.5281/zenodo.19603799) and the public GitHub repository linked from the notebook [16]. The notebook prints the runtime software versions at launch; the validation environment used for the results reported here ran PyTorch 2.10.0+cu128 and Matplotlib 3.10.0, and each execution records the exact versions present in that environment. PyTorch and Matplotlib are cited here as the software frameworks used in the study [17,18].

## 2.2. Gray–Scott substrate

The substrate evolves two fields, U and V, on a $96 \times 96$ grid with time step $\Delta t = 1.0$ and diffusion coefficients $D_u = 0.16$ and $D_v = 0.08$. The governing equations are $\partial_t U = D_u \nabla^2 U - UV^2 + F(1 - U)$ and $\partial_t V = D_v\nabla^2 V + UV^2 - (F + K)V$. We discretize the Laplacian with a 3 × 3 isotropic stencil [[0.05, 0.20, 0.05], [0.20, −1.00, 0.20], [0.05, 0.20, 0.05]] and reflective padding, which approximates a no-flux tissue boundary. This choice makes the distributed nature of the substrate explicit: each update uses only the local neighbourhood, and no centralized state is broadcast across the field. This term also shapes the dynamical system's pattern selection: the Laplacian diffuses local differences, damps pixel-scale noise and couples each site to wavelength-dependent spatial modes whose growth or decay depends on the reaction terms. In the Gray–Scott system, that interaction is what allows small perturbations to be amplified into stable ring- and spot-like Turing textures rather than remaining as unstructured noise.

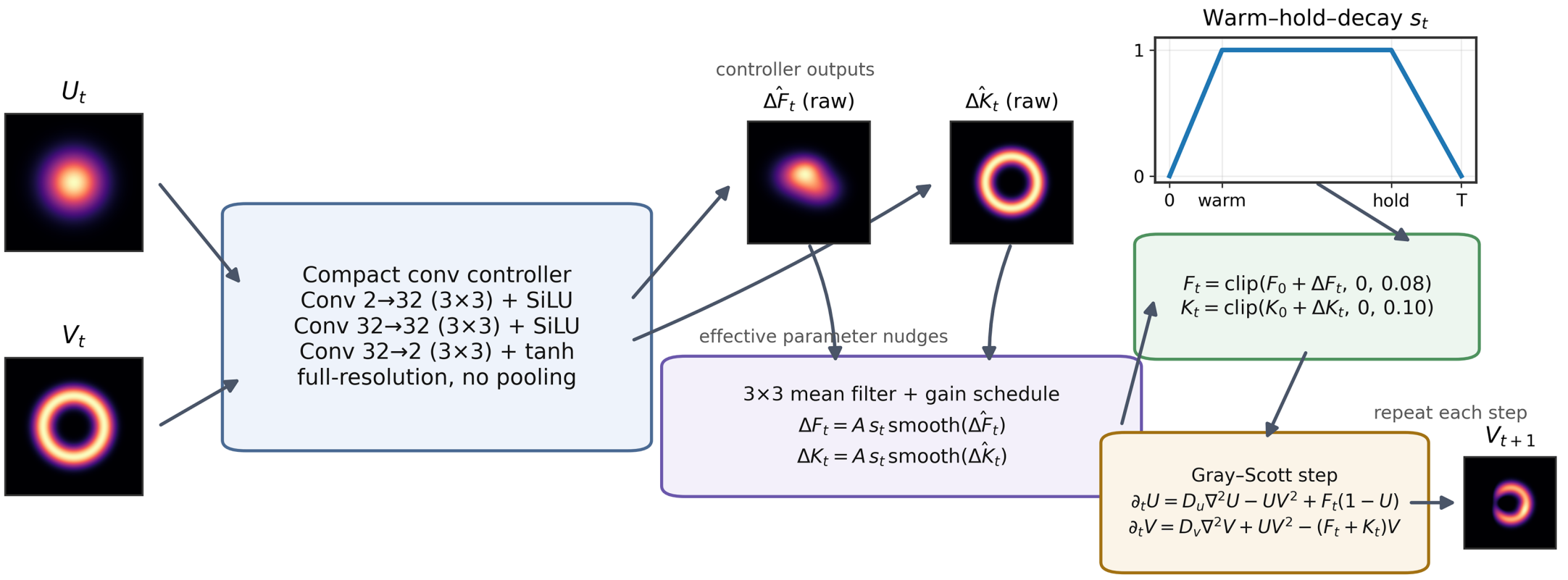

**Figure 1.** Hybrid controller overview and parameter-level nudges. The controller observes the full U and V fields, outputs smooth ΔF and ΔK maps and acts only through parameter modulation under a warm-hold-decay schedule. The figure is schematic: it clarifies what the controller can influence and what it cannot.

Initial conditions set U ≈ 1 and V ≈ 0 with a small central V seed and additive Gaussian noise to break symmetry. The pure RD baseline uses $(F_0, K_0) = (0.04, 0.06)$ and zero control amplitude. The NN-dominant baseline weakens the substrate to (0.01, 0.01) and uses high-amplitude control. The hybrid regime keeps the self-organizing substrate at (0.04, 0.06) and uses a moderate control amplitude.

## 2.3. Controller architecture

The controller is compact and fully convolutional. It observes both U and V at full spatial resolution; there is no pooling and no down-sampling. Concretely, the network applies Conv(2→32, 3 × 3), sigmoid linear unit (SiLU), Conv(32→32, 3 × 3), SiLU and Conv(32→2, 3 × 3), followed by tanh to keep the raw outputs in [−1, 1]. The two output channels are interpreted as raw parameter proposals for ΔF and ΔK. A 3 × 3 mean filter smooths these outputs to encourage field-level modulations rather than pixel-scale actuation.

## 2.4. Action parametrization and schedule

The effective control fields are applied after smoothing and after the warm-hold-decay gain schedule: $\Delta F_t = A\, s_t\, \mathrm{smooth}(\Delta\hat{F}_t)$ and $\Delta K_t = A\, s_t\, \mathrm{smooth}(\Delta\hat{K}_t)$. The partial differential equation (PDE) then uses $F_t = \mathrm{clip}(F_0 + \Delta F_t, 0, 0.08)$ and $K_t = \mathrm{clip}(K_0 + \Delta K_t, 0, 0.10)$. The reported L1 effort and L2 power are computed on these final effective modulations after the gain schedule has been applied, not on the raw network outputs. In the code, this is implemented as the per-step spatial mean of $|\Delta F_t| + |\Delta K_t|$ for L1 and $\Delta F_t^2 + \Delta K_t^2$ for L2, averaged across time. This controller-substrate workflow is summarized in figure 1.

## 2.5. Spectral readout

Pattern quality is measured from the fast Fourier transform (FFT) power spectrum of V after mean removal. Let $P_t(f)$ denote the two-dimensional spectral power at time t. The target annulus is the radial band $0.05 \leq ||f|| \leq 0.22$. Band selectivity is $\mathrm{band_ratio}(V_t) = \Sigma_{f \in \mathrm{band}} P_t(f) / \Sigma_{f \neq 0} P_t(f)$, and absolute band power is $\mathrm{band_abs}(V_t) = (1/HW)\, \Sigma_{f \in \mathrm{band}} P_t(f)$. These quantities reward Turing-like ring structure while suppressing trivial solutions that merely generate weak or spectrally diffuse patterns.

**Table 1.** Key outcomes by regime (means over $n = 8$ seeds in the full-horizon evaluation). Here, stability_tail is a descriptive final-tail summary rather than the earliest qualifying hold-window value used to declare strict convergence at t*; it may therefore exceed $1 \times 10^{-5}$ at the end of the rollout even when the strict-convergence criterion was satisfied earlier.

| regime | conv. (strict/quasi) | t* strict (steps) | final band ratio | abs. band power | L1 effort | L2 power | stability_tail |
|---|---|---|---|---|---|---|---|
| pure RD | 0%/0% | 240 | 0.434 | $7.146 \times 10^{-4}$ | 0 | 0 | $1.544 \times 10^{-5}$ |
| NN-dominant | 0%/0% | 240 | 0.077 | $3.106 \times 10^{-3}$ | $1.183 \times 10^{-1}$ | $8.258 \times 10^{-3}$ | $6.464 \times 10^{-4}$ |
| hybrid | 100%/100% | 165 | 0.438 | $6.454 \times 10^{-4}$ | $7.902 \times 10^{-3}$ | $4.078 \times 10^{-5}$ | $2.217 \times 10^{-5}$ |

## 2.6. Loss function

The training objective is $\mathscr{L}$ = w_band L_band + w_abs L_abs + w_stab L_stab + w_sustain L_sustain + w_ctrl L_ctrl + w_TV L_TV, with (w_band, w_abs, w_stab, w_sustain, w_ctrl, w_TV) = (0.5, 0.9, 0.8, 0.9, 0.08, 0.0).

L_band = 1 − band_ratio (V_T), L_abs = max (0, 1 − band_abs (V_T) / ($1.2 \times 10^{-4}$)).

Let d_t = (1/HW) $\|$V_t − V_{t−1}$\|_1$, and let đ_tail be the mean of d_t over the last CONV_WIN = 5 steps. With spectral gate g_T = $\sigma$ (12 · min ((band_ratio (V_T) − 0.22)/0.22, (band_abs (V_T) − $1.2 \times 10^{-4}$)/($1.2 \times 10^{-4}$))), the stability term is L_stab = g_T · max (0, (đ_tail – $10^{-5}$)/($10^{-5}$)).

For the final SUSTAIN_TAIL = 24 rollout steps, L_sustain = mean_t [max (0, (0.22 − band_ratio (V_t))/0.22) + max (0, ($1.2 \times 10^{-4}$ – band_abs (V_t))/($1.2 \times 10^{-4}$))].

The control term is L_ctrl = (1/T) Σ_t mean (|ΔF_t| + |ΔK_t|), computed on the effective modulations after smoothing and gain scheduling. The optional spatial total-variation term L_TV was set to zero in the reported experiments.

## 2.7. Convergence definitions

Let d_t = (1/HW) $\|$V_t – V_{t−1}$\|_1$ and let mov_t denote its five-step moving average. Strict convergence is the earliest time t* such that, over a 12-step hold window ending at t*, mov_t < $1 \times 10^{-5}$, band_ratio (V_t) ≥ 0.22 and band_abs (V_t) ≥ $1.2 \times 10^{-4}$ for every step in the window. Quasi-convergence uses the same stability and spectral gates, and additionally requires plateau stability within the hold window: max_t |b_t – b̄| / |b̄| ≤ 0.05 for b_t = band_ratio (V_t) and max_t |a_t – ā| / |ā| ≤ 0.08 for a_t = band_abs (V_t). This distinction explains why the main regime comparison reports t* = 165 steps for strict convergence, whereas the Goldilocks amplitudes in the shorter sweep reach quasi-convergence in 94–96 steps.

## 2.8. Training and evaluation

Controllers are trained end-to-end with Adam, gradient clipping, deterministic seeds and a long-horizon curriculum. Table 1 summarizes the full-horizon regime comparison as means over $n = 8$ seeds. Table 2 reports the sparse five-point hybrid amplitude sweep as seed-aggregated means over $n = 4$ seeds, and table S1 in the supplementary material reports the executed denser 11-point follow-up sweep over the same trained hybrid controller weights, also with $n = 4$ seeds and the same 120-step quasi-convergence horizon. Table 3 reports a targeted boundary/grid robustness scan, again with $n = 4$ seeds per condition. Table S2 in the supplementary material isolates the Goldilocks boundaries A = 0.030 and A = 0.045 and reports explicit success counts together with mean and observed s.d. of quasi-convergence time across the same $n = 4$ seeds. These sweep values are therefore not single illustrative rollouts but multi-seed summaries. In that boundary-specific check, both amplitudes quasi-converged in 4/4 seeds, and the observed standard deviation of t~quasi was 0.0. Tables S1 and S2 in the supplementary material and table 3 are included to show that the reported Goldilocks interval and cost ordering are not artefacts of one noise realization, although this robustness analysis is not exhaustive. Non-converged amplitudes are included in the sweep plots as infeasible points but excluded from the feasible-set interpretation itself.

**Table 2.** Hybrid amplitude sweep with absolute and relative control costs (seed-aggregated means over $n = 4$ seeds in the sparse sweep). Table S2 in the supplementary material reports explicit success counts and mean ± s.d. summaries for the two Goldilocks boundaries A = 0.030 and A = 0.045.

| AMP | quasi conv. (%) | t∼ quasi (steps) | band ratio | L1 effort | L2 power | L2 (rel. A = 0.030) |
|---|---|---|---|---|---|---|
| 0.000 | 0 | 120 | 0.473 | 0 | 0 | 0.00× |
| 0.015 | 0 | 120 | 0.383 | $3.957 \times 10^{-3}$ | $1.026 \times 10^{-5}$ | 0.25× |
| 0.030 | 100 | 96 | 0.296 | $7.916 \times 10^{-3}$ | $4.104 \times 10^{-5}$ | 1.00× |
| 0.045 | 100 | 94 | 0.277 | $1.187 \times 10^{-2}$ | $9.222 \times 10^{-5}$ | 2.25× |
| 0.080 | 0 | 120 | 0.382 | $2.105 \times 10^{-2}$ | $2.889 \times 10^{-4}$ | 7.04× |

**Table 3.** Boundary/grid robustness scan (seed-aggregated means over $n = 4$ seeds per setting).

| setting | quasi (%) | t∼ | band | L1 | L2 |
|---|---|---|---|---|---|
| 96 refl., A = 0.030 | 100 | 96.0 | 0.296 | $7.916 \times 10^{-3}$ | $4.10 \times 10^{-5}$ |
| 96 refl., A = 0.045 | 100 | 94.0 | 0.277 | $1.187 \times 10^{-2}$ | $9.22 \times 10^{-5}$ |
| 96 circ., A = 0.030 | 100 | 96.0 | 0.296 | $7.890 \times 10^{-3}$ | $4.10 \times 10^{-5}$ |
| 96 circ., A = 0.045 | 100 | 94.0 | 0.277 | $1.183 \times 10^{-2}$ | $9.22 \times 10^{-5}$ |
| 128 refl., A = 0.030 | 100 | 84.5 | 0.285 | $8.045 \times 10^{-3}$ | $4.20 \times 10^{-5}$ |
| 128 refl., A = 0.045 | 100 | 79.0 | 0.265 | $1.207 \times 10^{-2}$ | $9.40 \times 10^{-5}$ |

### 2.9. Pareto front and practical knee

Figure 2a shows the sparse five-point sweep used in the main manuscript, but the feasible-set interpretation is based only on quasi-converged amplitudes. Formally, the empirical Pareto front in this manuscript is the subset of quasi-converged feasible amplitudes for which no other feasible point attains both lower control cost and higher band selectivity. In that sparse sweep A = 0.030 and A = 0.045 are both feasible, yet A = 0.045 is dominated because it attains lower band selectivity at higher control cost. Table S1 in the supplementary material reports the denser 11-point follow-up sweep; it adds intermediate feasible amplitudes A = 0.035 and A = 0.040, but neither improves on A = 0.030 in the joint sense of lower cost and higher band selectivity. We therefore describe A = 0.030 as the practical knee because it is the lowest-cost feasible point and, in the executed sweeps, the only non-dominated feasible point, while acknowledging that the denser scan refines the feasible interval rather than overturning the sparse-sweep interpretation.

## 3. Results

### 3.1. Regime comparison

Table 1 summarizes the three main regimes. The hybrid model is the only one that converges reliably within the evaluation horizon: it reaches 100% strict and 100% quasi-convergence, with mean strict convergence time t* = 165 steps. Both pure RD and the NN-dominant baseline fail to converge within 240 steps under the same detector. Representative pattern evolution across the three regimes is shown in figure 3. This addresses the key question of division of labour directly: self-organization alone yields the right kind of texture but not the required stationarity, whereas strong centralized control expends large effort without producing the correct spectral structure.

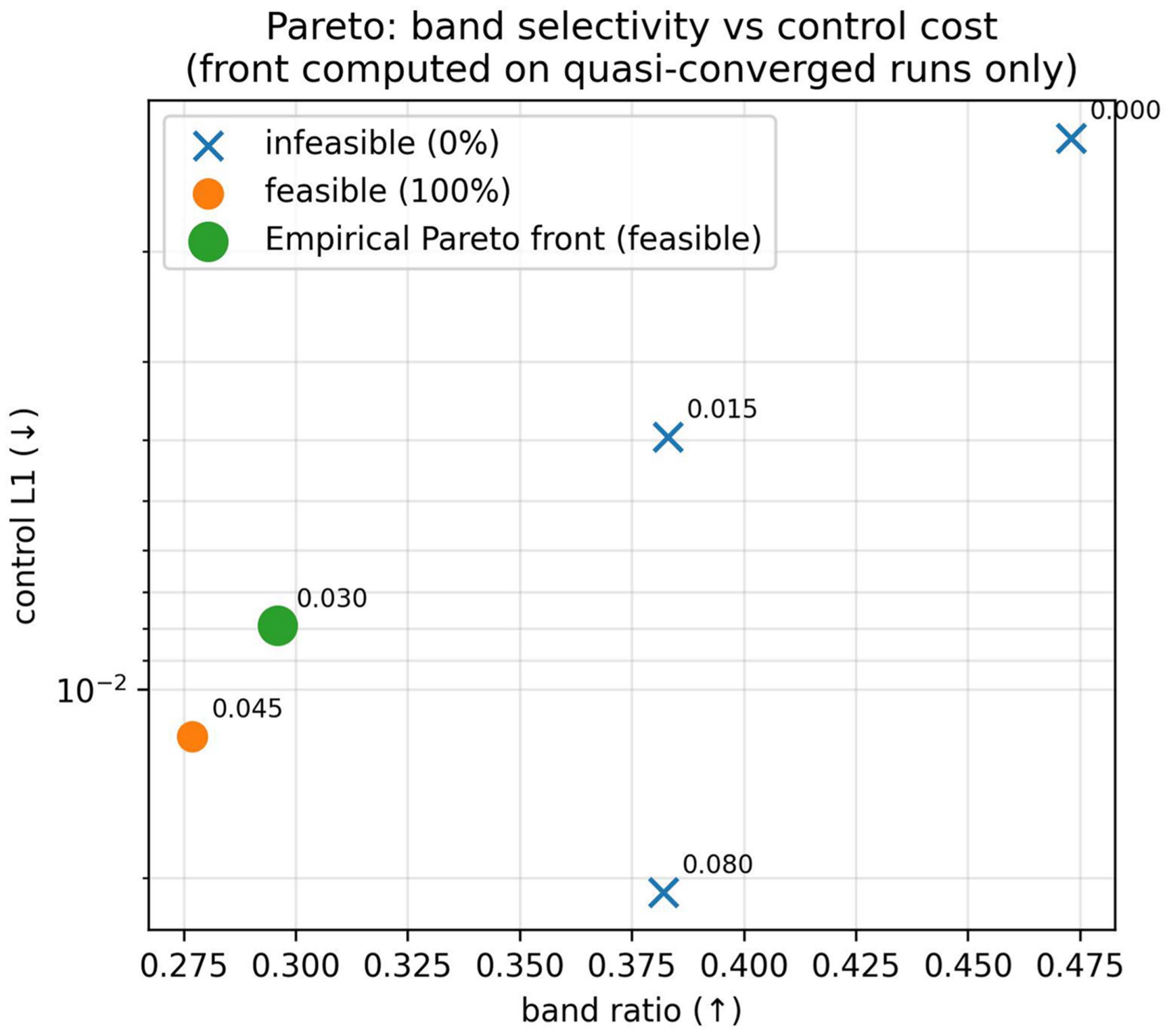

(a)

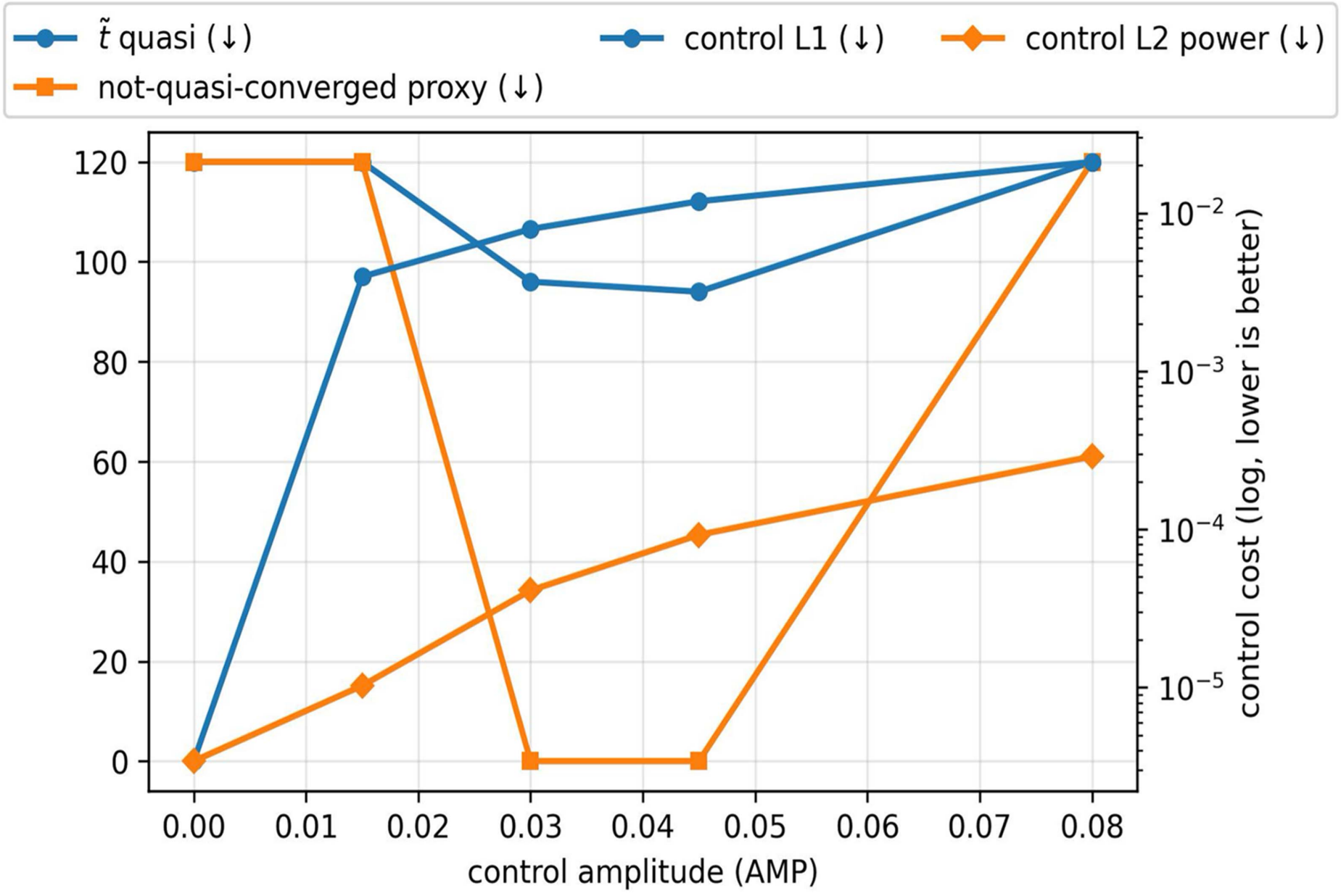

(b)

**Figure 2.** (a) Pareto trade-off: band selectivity versus control cost. Each point is the mean over $n = 4$ seeds from the sparse sweep. All five amplitudes are shown, but the empirical Pareto front is defined only on quasi-converged feasible amplitudes; a point is non-dominated if no other feasible point attains both lower cost and higher band selectivity. In this sparse sweep, A = 0.030 is the only non-dominated feasible point, while A = 0.045 remains feasible but dominated. Table S1 in the supplementary material shows that the denser 11-point follow-up sweep preserves the same practical-knee interpretation. (b) Division of labour versus control amplitude. Moderate amplitudes define a Goldilocks zone with fast quasi-convergence and low controller cost, while both weaker and stronger control fail within the 120-step sweep horizon. Points are means over $n = 4$ seeds from the sparse sweep. Table S1 in the supplementary material sharpens the feasible interval to A = 0.030–0.045 in the tested 11-point follow-up sweep.

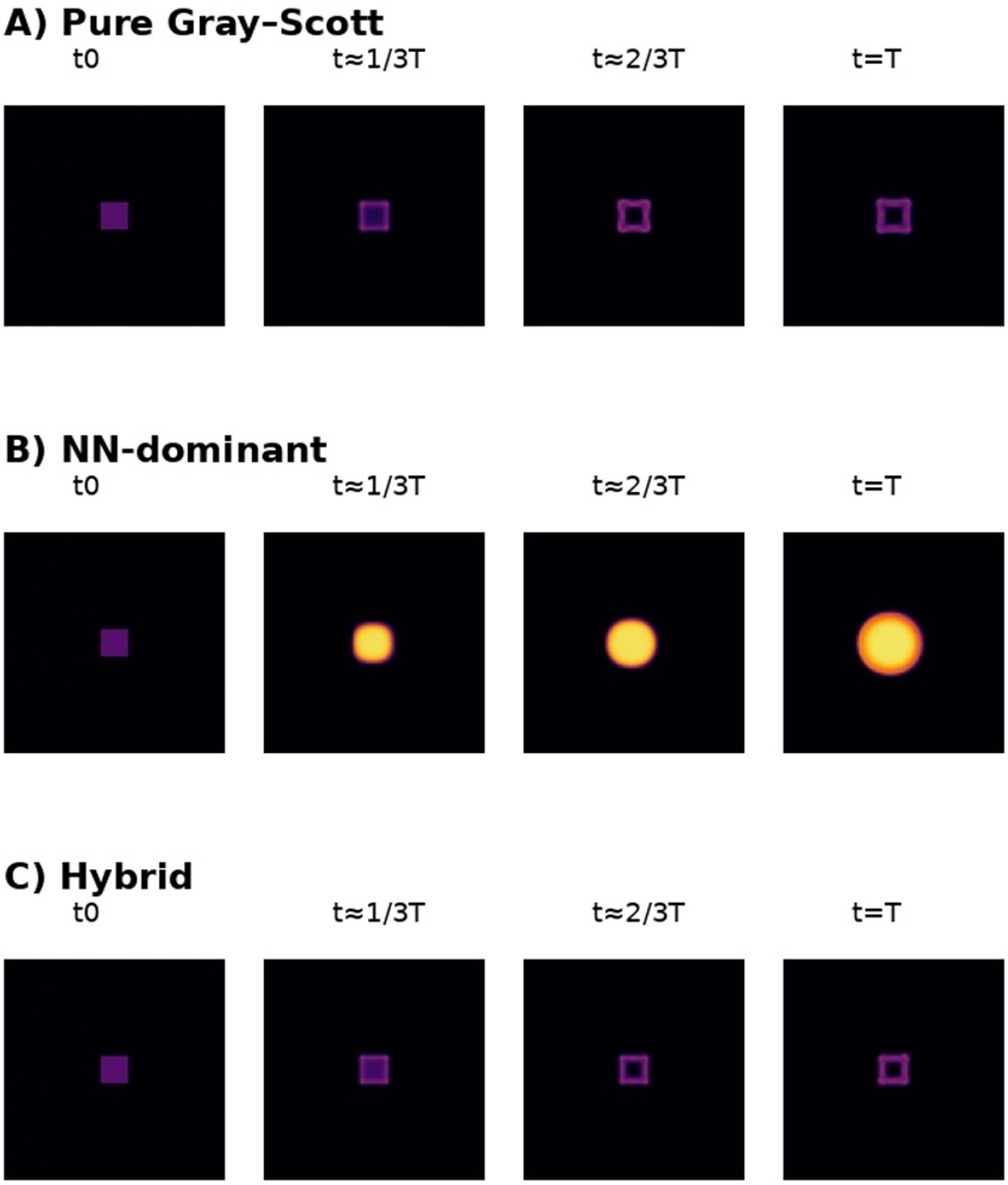

**Figure 3.** Representative pattern evolution across regimes. Pure RD and hybrid end in visually similar ring-like textures, but the NN-dominant baseline produces a spectrally incorrect high-power blob. The main difference between pure RD and hybrid is therefore temporal and stability-related, not merely visual.

The hybrid controller also achieves the best quality-cost trade-off. Its final band ratio (0.438) closely matches the self-organized Pure-RD pattern (0.434), but its control usage is extremely low relative to the NN-dominant baseline. Comparing absolute costs, the hybrid uses roughly 15× less L1 effort ($7.902 \times 10^{-3}$ versus $1.183 \times 10^{-1}$) and more than 200× less L2 power ($4.078 \times 10^{-5}$ versus $8.258 \times 10^{-3}$). This is the central empirical signature of seed then cede: the controller does not paint the final pattern directly; it supplies small early corrections and then gets out of the way.

## 3.2. Goldilocks amplitudes

Table 2 reports the sparse five-point hybrid amplitude sweep as seed-aggregated means over $n = 4$ seeds, using absolute L1 and L2 costs, with the previously reported relative L2 multiplier retained as an extra column. The moderate amplitudes A = 0.030 and A = 0.045 achieve 100% quasi-convergence in 96 and 94 steps, respectively, while lower and higher amplitudes fail within the 120-step sweep horizon. These 94–96-step values are therefore aggregated threshold-crossing times rather than single representative rollouts. Table S2 in the supplementary material makes the boundary statistics explicit: A = 0.030 and A = 0.045 each quasi-converged in 4/4 seeds, with t~quasi = 96.0 ± 0.0 and 94.0 ± 0.0 steps, respectively. Within this $n = 4$ boundary check, there was no observed seed-to-seed variation in quasi-convergence success or threshold-crossing time. Table S1 in the supplementary material extends this analysis to 11 amplitudes and shows that the feasible set in the tested range is A = 0.030, 0.035, 0.040, and A = 0.045, with mean quasi-convergence times 96.0, 95.25, 95.0 and 94.0 steps. The left and right feasibility boundaries in the executed follow-up sweep, therefore, lie between A = 0.025/0.030 and A = 0.045/0.055. The key interpretive point remains that the manuscript's headline 165-step value refers to strict convergence in the main three-regime comparison, whereas the 94–96-step values refer to quasi-convergence in the shorter amplitude scans.

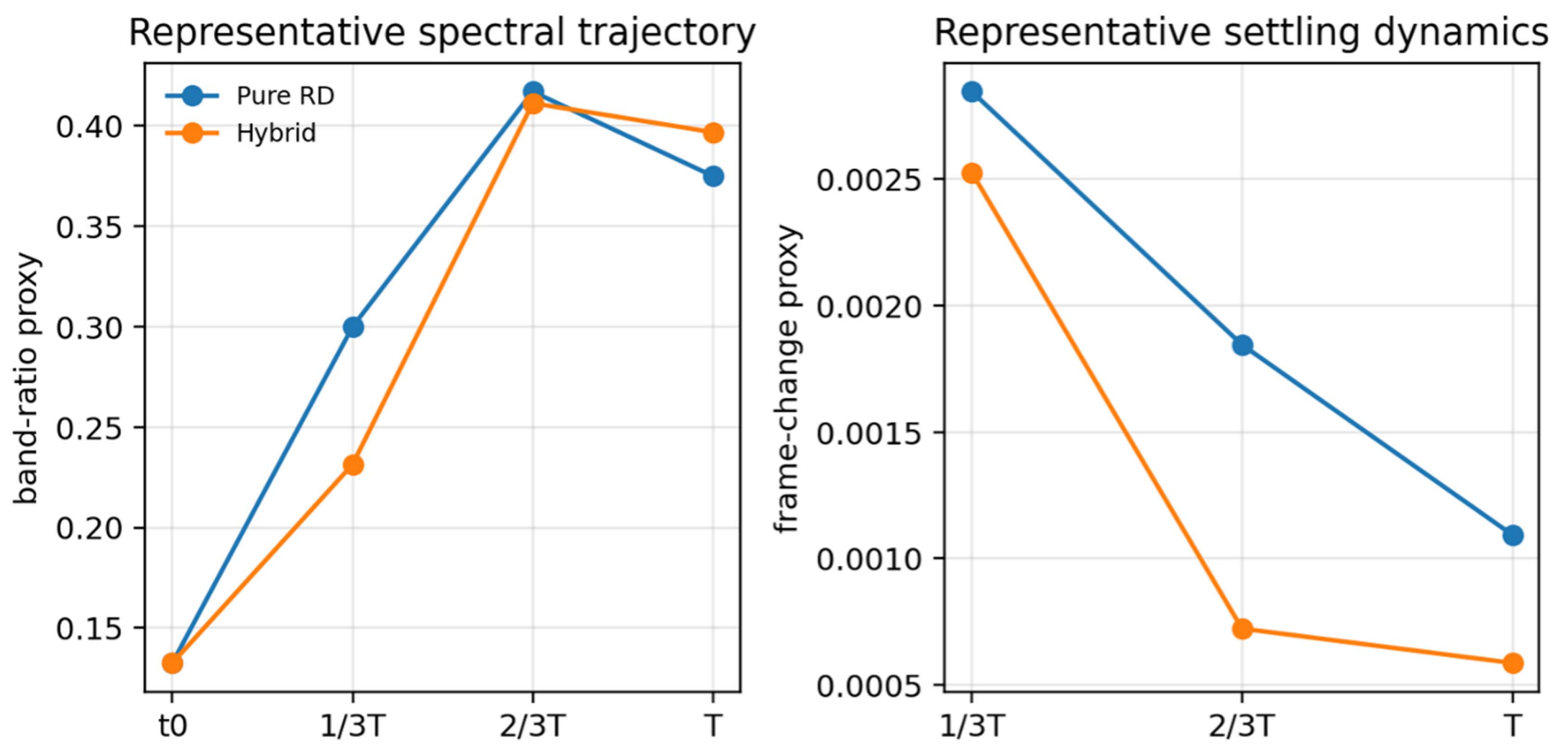

**Figure 4.** Representative temporal diagnostics derived from the displayed trajectories in figure 3. Left: an image-space spectral proxy. Right: frame-to-frame change proxy. These proxies show earlier settling in the hybrid rollout. Figure S1 in the supplementary material reports exact raw-field diagnostics derived from band_ratio (V_t), band_abs (V_t) and d_t = (1/HW) ||V_t − V_{t−1}||$_1$ for representative pure-RD and hybrid rollouts.

Figure 2a plots band selectivity versus control cost for the sparse sweep shown in table 2. All five amplitudes are shown, but the feasible-set interpretation is restricted to the quasi-converged amplitudes. In that sparse view, A = 0.030 and A = 0.045 are both feasible, but A = 0.045 is dominated by A = 0.030 because it has lower band selectivity at higher control cost. Table S1 in the supplementary material shows that the same dominance relation persists when the intermediate amplitudes A = 0.035 and A = 0.040 are added. Accordingly, A = 0.030 remains the practical knee/lowest-cost feasible point. Figure 2b shows the sparse scan as a function of amplitude, making the non-monotonic Goldilocks zone visually explicit.

## 3.3. Temporal dynamics

The final pure-RD and hybrid snapshots can appear visually similar at the final time step, so figure 4 reports a representative temporal comparison. The hybrid trajectory settles earlier than pure RD: the spectral proxy reaches the target band while the frame-to-frame change proxy collapses faster, consistent with the interpretation that the main difference between the two regimes is temporal and stability-related rather than purely visual. Figure S1 in the supplementary material provides exact raw-field diagnostics for representative rollouts, including band_ratio (V_t), band_abs (V_t) and the five-step moving average of d_t.

## 3.4. Boundary/grid robustness

A small multi-seed robustness scan summarized in table 3 examines sensitivity to boundary conditions and grid size. At the baseline 96 × 96 grid, replacing reflective padding with periodic/circular padding leaves the moderate-amplitude results essentially unchanged: both A = 0.030 and A = 0.045 remain 100% quasi-converged with the same t~ values (96 and 94 steps). On a larger 128 × 128 reflective grid, both amplitudes again remain 100% quasi-converged and in fact settle faster (84.5 and 79.0 steps). Across all scanned settings, A = 0.030 remains the lower-cost feasible point. These checks are limited in scope but, together with table 2 and table S1 in the supplementary material, they support the claim that the moderate-gain hybrid advantage is not an artefact of one boundary choice or one noise draw.

# 4. Discussion

These results sharpen the study's conceptual claim. The controller is not presented as a full model of cognition in the strong philosophical or neurobiological sense. Rather, the present framework is a minimal operational model for measuring the division of labour between centralized guidance and distributed self-organization: a centralized policy scaffolds a distributed pattern-forming substrate, and the resulting trade-off can be quantified directly. This is why the paper uses brain-like and cell-like as operational metaphors and explicitly states that the substrate is distributed because it evolves through local reaction and diffusion updates, not because the Gray–Scott variables literally represent biological cells.

This operational stance also clarifies the connection to positional information and developmental control. A centralized policy that reads the whole field and writes smooth parameter-level biases is one possible abstraction of a higher-level instructive signal. The substrate then performs the detailed spatial computation. That is why the hybrid controller can be interpreted as providing scaffolding rather than direct synthesis: it biases the dynamics towards a productive basin of attraction and then lets the local RD process finish the job [8].

A useful biological analogy comes from shell morphogenesis. Boettiger, Ermentrout and Oster argued that shell structure and shell pigmentation can reflect a neural or neurosecretory influence on an underlying local patterning process [9]. Our model is much more abstract than that of biology, but it captures the same qualitative possibility: global guidance can bias local self-organization without explicitly specifying every final feature. Likewise, recent work on sensorimotor cellular automata shows that agency-like behaviour can emerge from local rule systems coupled to training signals [10,11]. Compared with those models, the present study uses a PDE substrate rather than a learned local update rule and focuses specifically on parameter-level guidance of morphogenesis.

A natural methodological question is why a convolutional controller is used instead of reinforcement learning, a genetic algorithm or a classical controller. The answer is practical and structural. A compact convolutional neural network (CNN) supplies the right inductive bias for smooth spatial fields, supports differentiable end-to-end training with the FFT objective, and makes it easy to constrain the controller to parameter-level nudges rather than state overwriting. That does not imply that CNN control is uniquely correct. A matched comparison against reinforcement learning (RL) or classical control remains important future work and is stated explicitly in the manuscript's limitations.

Applications remain a major motivation. In soft robotics, the lesson is not simply to control less, but to control at the right moment and at the right level of abstraction. In bioengineering and self-assembly, the same pattern suggests that small instructive cues may be more scalable than strong global forcing. The empirical message is therefore broader than Gray–Scott itself: when the substrate is rich enough, good control often means steering the attractor landscape rather than commanding every detail.

# 5. Conclusion

The central result of this study is that within the tested regimes, reliable morphogenesis is achieved most efficiently by a hybrid controller–substrate partnership rather than by either self-organization alone or centralized control alone. The hybrid regime attains 100% strict convergence at approximately 165 steps and reaches the target texture with far lower effort than the NN-dominant baseline.

The framework specifies the exact loss, the exact convergence detectors, the exact meaning of the L1/L2 costs, the full controller architecture, the seed aggregation used in each table and the practical interpretation of the Pareto knee across both the sparse manuscript sweep and the denser 11-point follow-up sweep. The robustness scan further shows that this qualitative result survives a periodic boundary at $96 \times 96$ and a larger $128 \times 128$ reflective grid. These details strengthen the paper's main claim: under the tested conditions, the most effective policy is to seed then cede, using brief smooth parameter-level nudges to put the substrate on the right trajectory and then letting local dynamics do the rest.

Viewed more broadly, the study offers a concrete quantitative route to morphological computation. The division of labour studied here should be read as an operational and measurable trade-off between centralized guidance and distributed self-organization, not as a strong philosophical claim that the controller itself instantiates cognition. That makes the hybrid framework useful both as a scientific probe of embodied morphogenesis and as a design principle for systems that compute through the coordinated action of controller and material substrate.

# 6. Limitations

The primary reported benchmark remains the $96 \times 96$ reflective-boundary Gray–Scott system. A targeted multi-seed robustness scan shows that switching to periodic/circular boundaries on the $96 \times 96$ grid leaves the moderate-gain results essentially unchanged, and that a $128 \times 128$ reflective grid still preserves 100% quasi-convergence for A = 0.030 and A = 0.045 while settling faster. These seed-aggregated checks reduce concern that the result is a single-boundary or single-seed artefact, but they are not exhaustive enough to justify a broad claim of geometry-independent generality.

The main display in the manuscript remains sparse for readability. Table S1 in the supplementary material now reports the executed denser 11-point follow-up sweep, which confirms the same Goldilocks zone and places the feasibility boundaries between A = 0.025/0.030 and A = 0.045/0.055 in the tested set, with A = 0.030 remaining the lowest-cost feasible point. Table S2 in the supplementary material further shows that the two boundary amplitudes A = 0.030 and A = 0.045 each quasi-converged in 4/4 sampled seeds with observed t~quasi s.d. = 0.0. Even so, these boundary-specific variance checks are still based on $n = 4$ seeds and should be read as supportive rather than as a high-powered uncertainty analysis. The amplitude axis is only one-dimensional and sampled coarsely enough that it should not be over-interpreted as an exact global optimum. A more exhaustive sweep or multi-objective search would still be needed for a fully precise Pareto-front estimate.

The controller is deliberately minimal: memoryless, convolutional and restricted to smooth parameter-level modulation. This design is appropriate for studying minimal intervention, but it excludes richer controllers and does not benchmark against matched RL or classical control baselines.

Finally, the paper remains a computational study. Real developmental or robotic systems introduce partial observability, delays, actuator limits, model mismatch and experimental noise. Establishing whether the same minimal-intervention principle survives those conditions remains a key next step.

**Ethics.** This work did not require ethical approval from a human subject or animal welfare committee.

**Data accessibility.** All results reported in this article derive from numerical simulations of the Gray–Scott RD system. The accompanying notebook and the archived publication package are available through Zenodo (takeisika/morphological-computation: release for publication [16]) and the linked public GitHub repository.

**Declaration of AI use.** As a non-native English speaker, I used generative AI tools (Gemini and ChatGPT) to improve the grammar, clarity and phrasing of the manuscript, and to obtain suggestions for coding. All AI-assisted outputs were carefully reviewed, edited and validated by the author, who takes full responsibility for the final text, code and scientific content. No AI tool is listed as an author.

**Authors' contributions.** T.I.: conceptualization, data curation, formal analysis, investigation, methodology, resources, software, validation, visualization, writing—original draft, writing—review and editing.

**Conflict of interest declaration.** I declare I have no competing interests.

**Funding.** No funding has been received for this article.